\documentclass[journal]{IEEEtran}
\usepackage{filecontents}
\usepackage{textcomp}

\usepackage{cite}

\usepackage{graphicx}
\newcommand{\figref}[1]{\figurename~\ref{#1}}
\usepackage{amsmath}
\usepackage{url}

\begin{document}
%
\title{Preprocessing Methods and Pipelines of Data Mining: An Overview}
%
%
%

\author{
    \IEEEauthorblockN{Li Canchen} \\
    \IEEEauthorblockA{
    Department of Informatics \\
    Technical University of Munich \\
    Canchen.Li@in.tum.de}
}

%
%

\markboth{Seminar Data Mining, June~2019}%
{}
%



\maketitle

\begin{abstract}
Data mining is about obtaining new knowledge from existing datasets. However, the data in the existing datasets can be scattered, noisy, and even incomplete. Although lots of effort is spent on developing or fine-tuning data mining models to make them more robust to the noise of the input data, their qualities still strongly depend on the quality of it. The article starts with an overview of the data mining pipeline, where the procedures in a data mining task are briefly introduced. Then an overview of the data preprocessing techniques which are categorized as the data cleaning, data transformation and data preprocessing is given. Detailed preprocessing methods, as well as their influenced on the data mining models, are covered in this article. 
\end{abstract}


%
\IEEEpeerreviewmaketitle

\section{Introduction}
%
%
%
%

\IEEEPARstart{D}{ata} mining is a knowledge obtaining process: it gets data from various data sources and finally transforms the data into knowledge, thus provides insight to its application field. Data mining pipeline is a typical example of the end-to-end data mining system: they are an integration of all data mining procedures and deliver the knowledge directly from data source to human.

The purpose of data preprocessing is making the data easier for data mining models to tackle. The quality of data can have a significant influence on data mining models. It is considered that the data and features have already set the upper bound of the knowledge that can be obtained, and the data mining models are just about approximating the upper bound. Various preprocessing techniques are invented to make the data meet the input requirements of the model, improve the relevance of the prediction target, and make the optimization step of the model easier.

It is common that raw data obtained from the natural world is badly shaped. The problems include the appearance of missing values (e.g., a patient did not go through all the tests), duplications (e.g., annual income and monthly income), outlier values (e.g., age is -1) as well as contradictions (e.g., gender is male and is pregnant) in the dataset. Although the existing preprocessing techniques would not guarantee to solve all these problems, they could at least correct some of them and improve the performance of the models.

The data type and distribution of data are usually transformed before being sent to data mining models. The purpose of data transformation includes making the data meets the input requirement of the models, removing the noise of data, and making the distribution of data more suitable for applying optimization algorithms in the model training step.

The input for data mining models can be huge: they may have too many dimensions or of massive amount, which would make it difficult for the data mining model to train or cause troubles while transferring and storing the data. Data reduction techniques can reduce the problem by applying reduction on dimensions (known as dimensional reduction) or amounts of data (known as instance selection and sampling).

To implement preprocessing to data, Python and R are among the most popular tools. With bulks of packages such as scikit-learn\cite{scikit-learn} and PreProcess\cite{preprocess},  most of the preprocessing algorithms covered in this paper can be implemented even without consideration of its details. 

In the following section, the data mining pipeline and the primary procedures in the data mining pipeline will be introduced. From Section 2 on, we will focus on the steps in the data preprocessing work: Section 3 will introduce the techniques used in data cleaning, while Section 4 will cover the data transformation techniques. In the last section, data reduction techniques will be discussed.

\section{Data Mining Pipeline}

\begin{figure}[!t]
\centering
\framebox{\includegraphics[width=2.5in]{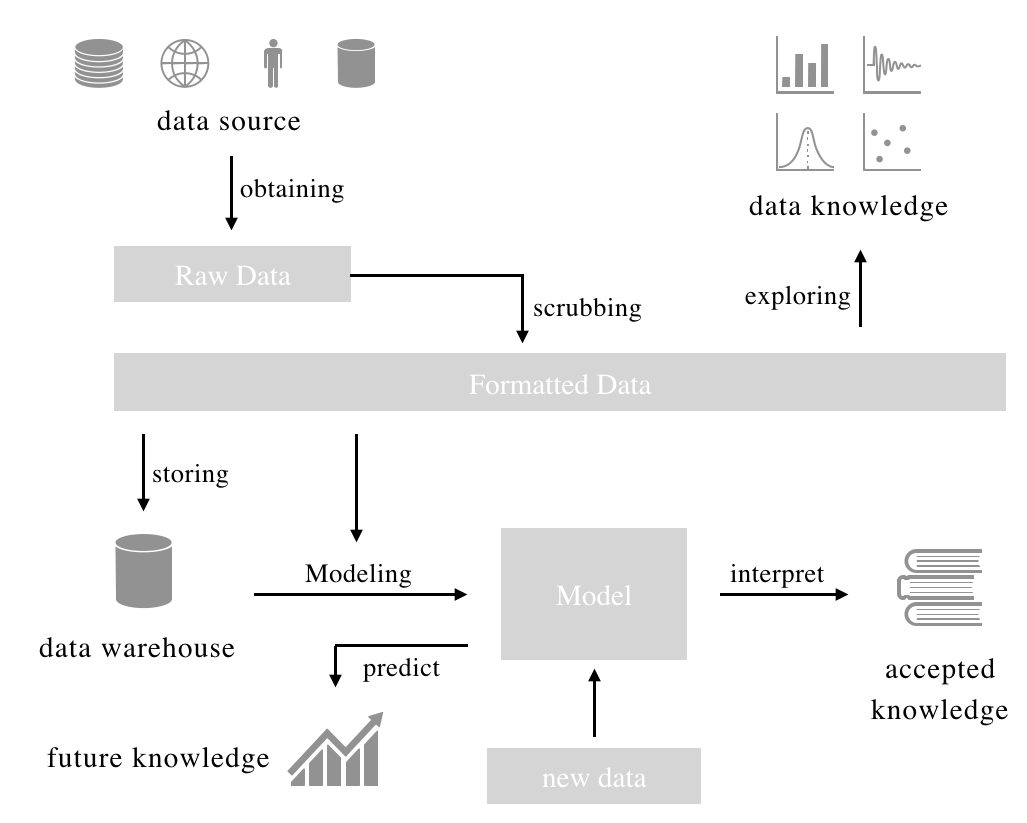}}
\caption{An illustration of data mining pipeline.}
\label{fig:datamining}
\end{figure}

The data mining pipeline is an integration of all procedures in a data mining task. While most of the data already exists in a data base, a data warehouse, or other types of data source\cite{han2011data}, various steps should be taken in order to make them easier for a human to understand. An illustration of the data mining pipeline is given as in \figref{fig:datamining}. Generally speaking, the key procedures include \textit{obtaining}, \textit{scrubbing}, \textit{exploring}, \textit{modelling}, \textit{interpreting}. These procedures are known as "OSEMN"\footnote{http://www.dataists.com/2010/09/a-taxonomy-of-data-science/}. However, note that the pipeline is not a linear process in the real world, but a successive and long-lasting task. Methods in scrubbing and modeling procedures have to be tested and refined, the obtaining procedure may have to be adapted for different kinds of data sources, and the visualization and interpretation of the data may have to be adjusted for their audience, thus meet the audience's demand. In the rest of this section, details of these procedures will be discussed.

\subsection{Obtaining}
Obtaining the data is the most fundamental step in data mining since it is the data itself that decides what knowledge it may contain. Data base and data warehouse are among the primary sources of data, where the structured data can be fetched with query languages, usually SQL. The data warehouse is specially designed for organizing, understand, and making use of the data\cite{han2011data}: they are usually separated system from the operational database, having time-variant structure as well as structures that makes the subsequent analysis work easy, and most importantly, nonvolatile.

The obtained data can be archived as files and directly used for the subsequent procedures. They may also be reformatted and stored in a data base or a data warehouse, prepared for the data mining tasks in the future.

In the past, as well as in most common cases now, we regard the data obtaining step as the process for obtaining a dataset, regardless of how we obtained them. However, nowadays, new data is generating at an extreme speed: there are tons of data being created every second of every day. Some services, such as the public opinion monitoring and the recommendation system, do need the newly generated data: they have strong demand for being on time. In these circumstances, the concept of "stream" is comparatively more important than a dataset. A stream is a real-time representation of data. Under this concept, models and algorithms that can run online are developed\cite{gaber2005mining}. For the stream mining tasks, the goal of data obtaining is no longer obtaining a dataset, but a real-time input source.

\subsection{Scrubbing}
Scrubbing is about the cleaning and preprocessing of the data, aiming to make the data have a unified format and easy to be modeled. As for the detailed concepts and techniques in data scrubbing, the readers could refer to the following sections of this paper, since most of them will be covered in the overview of data preprocessing.

\subsection{Exploring}
Before modeling the data, people may want to get to know the underlying distribution of the data, the correlation between variables, and their correlation with the labels. Assumptions can be made in this step. For instance, people may assume smoking is highly correlated with lung cancer. These assumptions are important because they would provide indications to other procedures in a data mining task, including helping to choose a suitable model, and helping to justify your work in the interpreting of data.

Tools for exploring the data and verifying the assumptions are usually statistical analysis and data visualization. The statistical analysis gives us theoretical probabilities, known as significance level, of our assumption being incorrect, while data visualization tools, such as ggplot\cite{ggplot} and D3\cite{bostock2011d3}, give us the impressions about the distribution of the data, help people to verify their assumption conceptually. Also, new patterns that are ignored in the assumption step might be found in the visualization step.

\subsection{Modeling}
With underlying patterns existed in the data source, modeling makes it possible to represent the pattern explicitly with the data mining models. For a data mining task, modeling would usually split the data into the training set and test set thus could score the accuracy of the model on a relatively "new" dataset. If the model contains hyperparameters, such as the parameter $k$ in a K-Nearest Neighbor (KNN) model, a cross validation set will be created for obtaining the best set of the hyperparameter.

For most of the data mining models, loss functions are defined. Generally, a loss function will have lower value if the model performs well. Besides, it usually has special features such as convexity, which makes the gradient-based optimizing algorithm performs better. With trainable parameters, a model's training step is about adjusting its parameter so that it gains lower loss on its training data. The specific definition of loss function depends on the model itself and the task. Mean squared error ($\sum_{i=1}^n (\hat{y}_i - y_i^2$) for regression task and cross entropy ($-\sum_{i=1}^n [y_i\log \hat{y_i} + (1 - y_i)\log (1 - \hat{y_i})]$) for classification task are frequently accepted loss functions. 

There are too many kinds of data mining models existed; their tasks include clustering, classification, and regression. The complexity of the models also varies: simple models such as linear regression only have a few parameters, a small amount of data will make the training step converge, while complex models such as AlexNet have millions of parameters\cite{krizhevsky2012imagenet}, and their training also require huge dataset. However, complex does not mean better: the model should be decided according to the predicting target of the task, the dataset size, the data type, etc. Sometimes, it is necessary to run different models over one dataset and find the most suitable data mining model.

\subsection{Interpreting}
While previous steps are generally pure science, the interpreting step is more humanistic. Knowledge can be extracted from the input data, but it takes extra effort to convince people to accept this knowledge. Although complicated statistics and models make the work looks more professional, for laymen, graphs, tables and well explained accuracy make it easier to understand and accept. Besides, social skills such as story telling and emotion quotient are also important: this procedure is about the human, not the data.

\section{Data Cleaning}
The data obtained from the natural world is usually badly shaped. Some of the problems, such as outlier, may affect the data mining model and produce a biased result. For instance, the outlier may affect the K-Means clustering algorithm by substantially distort the distribution of data\cite{velmurugan2010computational}. Other problems, if not handled, will make it impossible for the data to be analyzed by models, such as the Not a Number (nan) values in a data vector. Data cleaning techniques,  including missing value handling and outlier detection, were issued to tackle the problem. They make the gathered data suitable as the input of the model.

\subsection{Missing Values Handling}
Missing values is a typical kind of data incompleteness of dataset. Most of the data mining models would not tolerate the missing values of its input data: these values can not be used for comparison, not available for categorizing, and can not be operated with arithmetic algorithms. Thus, it is necessary to handle the missing values before pushing the dataset to data mining models.
The easiest way to deal with missing value is to drop the entire sample. This method is effective if the proportion of missing value in a dataset is not significant, however, if the number of missing values is not suitable for ignoring, or the percentage of missing value for each attribute is different\cite{han2011data}, dropping the samples with missing value would reduce the amount of dataset dramatically, the information contained in the dropped samples is not made use of. Another way to deal with missing values is by filling them, and there are varies methods for finding the suitable value to fill the missing value, some of them are listed as follows.

\subsubsection{Use special value to represent missing}
Sometimes the missing value itself has some meaning. For instance, in a patient's medical report, the missing value for uric acid means the patient did not go through the renal function test. Thus, using a certain value such as -1 makes sense, for they can be operated like normal values while having special meaning in the dataset.

\subsubsection{Use attribute statistics to fill}
Statistics such as mean, median, or mode can be obtained from non-missing values in the missing value's corresponding attribute. It is said that for a skewed dataset, the median would be a better choice\cite{han2011data}. However, this technique does not take the sample's other non-missing attributes into account.

\subsubsection{Predicting the value with known attributes}
If we assume, there exists a correlation between attributes, filling the missing value can be modeled as a prediction problem: predicting the value with the non-missing attributes with other samples as training data. The predicting methods includes regression algorithms, decision trees\cite{han2011data}, and K-Means\cite{patil2010missing}.

\subsubsection{Assigning all possible values}
For categorical attributes, given an example $E$ with $m$ possible values for its missing value, then $E$ can be replaced with $m$ new examples $E_1, E_2, \dots, E_m$. This missing value filling technique assumes the missing attribute does not matter for the example. Thus the value can be anyone in its domain\cite{grzymala1991unknown}.

Comparison of different missing value filling techniques is done in \cite{grzymala2000comparison}. In the work, different missing value filling techniques are tested on ten datasets for running simple and extended classification methods. The conclusion shows C4.5 decision tree method performs the best, ignoring the samples with missing values and assigning all possible values also performed well, and filling the values with mode perform the worst. However, the performance of missing value filling techniques may differ because of the feature of the dataset. As a result, most of the techniques are worth trying for a data mining task.

\subsection{Outlier Detection}
Outlier refers to the data sample that has a massive distance to most of the other samples. Although the rare case does not necessarily mean wrong (e.g., age = 150), most outliers are caused by measurement error or wrong recording, thus ignoring a rarely appearing case would not harm a lot. Although some of the models are robust against outliers, outlier detection is still recommended in data preprocessing work.

Statistics-based outlier detection algorithms are among the most commonly used algorithms, which assume an underlying distribution of the data\cite{kantardzic2011data} and regard the data examples which corresponding probability density lower than a certain threshold as the outliers. As the underlying distribution is unknown for most cases, the normal distribution is a good substitute, and its parameter could be estimated by the mean value and standard deviation of the data. The Mahalanobis distance\cite{kantardzic2011data}, as in \eqref{eq:mahalanobis}, is a scale-irrelevant distance between two data samples. The outlier can be decided by comparing the Mahalanobis distance between each sample and the mean value of all samples. Box-plot, as another kind of statistics-based outlier detection technique, can give the graphical representation of outlier by plotting the lower quartile and upper quartile along with the median\cite{walfish2006review}. 

\begin{equation}\label{eq:mahalanobis}
    D_M(x, y) = \sqrt{(x - y)^T\Sigma^{-1}(x - y)} 
\end{equation}

Without making any assumption of the distribution of the data, the distance-based outlier detection algorithm can detect the outlier by analyzing the distance between every two samples, thus determine the outliers. Simple distance-based outlier detection algorithms are not suitable for a large dataset, since for $n$ samples with $m$ dimensions, their complexity is usually $O(n^2m)$\cite{kantardzic2011data}, and each computation requires scanning all the samples. However, an extended cell-based outlier detection algorithm is developed in\cite{knorr2000distance}, which guarantee linear complexity over the dataset volume and no more than three dataset scans. The experiment shows this algorithm is the best for the dataset with dimension less than 4.

Sometimes, with consideration of temporal and spatial locality, an outlier may not be a separate point, but a small cluster. Cluster-based outlier detection algorithms consider clusters with small size as outlier clusters and clean the dataset by removing the whole cluster\cite{ben2005outlier}\cite{duan2009cluster}.

\section{Data Transformation}

The representation of data in different attributes varies: some are categorical, while some are numerical. For categorical values, they can be nominal, binary or ordinal\cite{han2011data}, and for numerical data, they can also have different statistical features including mean values and standard deviations. However, not all kinds of data meet the requirement of data mining models. Also, the difference among data attributes may bring troubles for the subsequent optimization work of data mining models. Data transformation is about modifying the representation of data so that they are qualified to be the input for data mining models, as well as making the optimization algorithm of the data mining model easier to take effect. 
\subsection{Numeralization}

Categorical values widely exists in the natural world, some of the operations, such as calculating the entropy between groups, can be done directly over categorical data. However, most operations are not applicable to categorical data. Thus, categorical data is supposed to be encoded into numerical data, making it meets the requirements of the models. The following encoding techniques are adopted for numeralization.
\begin{itemize}
    \item \textit{One-Hot encoding}: Regard each possible value of the categorical data as a single dimension, and use $1$ for the dimension which the sample belongs to the category, otherwise $0$.
    \item \textit{Sequential encoding}: For each possible value of the categorical data, assign it with a unique and numerical index. This is implemented as a kind of word encoding, as in\cite{angelidis2018multiple}.
    \item \textit{Customized encoding}: Customized encoding is based on rules designed for a certain task. For instance, word2vec\cite{mikolov2013distributed} is an encoding that can turn a word into a $300$ dimensional vector, with consideration of the word's meaning.
\end{itemize}

Generally, one-hot encoding is suitable for categorical data with fewer possible values; if there are too many possible values, such as English words, the encoded dataset would be huge and sparse. Sequential encoding would not produce huge output, but the encoded data is not as easy to separate as one-hot encoded data. Customized encoding, if carefully designed, usually perform well over a certain kind of task, but for other tasks, the encoding should be redesigned, and its design can take lots of efforts.

\subsection{Discretization}
Discretization of data is applied sometimes to meet the requirement of input of models, such as Naive Bayes which require its input to be countable\cite{rish2001empirical}. Also, it can smooth the noise. Discretization of data does not necessarily make the data categorical, but make the continuous values countable. The discretization of data can be achieved with unsupervised learning methods such as putting data into equal-width or equal-frequency slots, known as \textit{binning}, or clustering. Some supervised learning methods such as \textit{decision tree} can also be used for discretization of data\cite{han2011data}.

\subsection{Normalization}
Since different attribute usually adopts a different unit system, their mean value and standard deviation are usually not identical. However, the numerical difference would make some of the attributes look more "important", while others are not\cite{han2011data}. This impression could cause trouble for some models; one of the typical ones is KNN: larger value would strongly affect the distance comparison, making the model mainly consider attributes tend to have larger numerical values. Besides, for neuron network models, the different unit system will also have a negative influence on gradient descent optimizing methods, forcing it to adopt a smaller learning rate. To tackle the problem mentioned above varies of normalization methods are issued, some of them are listed as follows.

\subsubsection{Min-max normalization}
Min-max normalization is used for mapping the attribute from its range $[lb, ub]$ to another range $[lb_{new}, ub_{new}]$; the target range is usually $[0, 1]$ or $[-1, 1]$ \cite{garcia2015data}. For a sample with value $v$, the normalized value $v'$ is given as in \eqref{eq:minmax}.

\begin{equation} \label{eq:minmax}
    v' = \frac{v - lb}{ub - lb} (ub_{new} - lb_{new}) + lb_{new}
\end{equation}

\subsubsection{Z-score normalization}
If the underlying range of an attribute is unknown or outlier exists, min-max normalization is not feasible or could be strongly affected\cite{garcia2015data}. Another normalization approach is to transform the data so that it would have $0$ as mean and $1$ as standard deviation. Given the mean $\mu$ and standard deviation $\sigma$ of the attribute, the transformation is represented as in \eqref{eq:zcore}.

\begin{equation} \label{eq:zcore}
    v' = \frac{v - \mu}{\sigma}
\end{equation}

Note that if $\mu$ and $\sigma$ are unknown, they can be substituted with the sample mean and standard deviation.

\subsubsection{Decimal scaling normalization}
An easier way to implement the normalization is to shift the floating point of the data so that each value in an attribute would have an absolute value less than $1$, the transformation is given as in \eqref{eq:decimal_scaling}.

\begin{equation} \label{eq:decimal_scaling}
    v' = \frac{v}{10^j}
\end{equation}{}

For some cases, different attributes have an identical or similar unit system, such as the preprocessing of RGB-colored imaged. In these cases, normalization is not necessary. However, if this is not guaranteed, normalization is still recommended for all data mining tasks.

\subsection{Numerical Transformations}
\begin{figure}[!t]
\centering
\framebox{\includegraphics[width=2.5in]{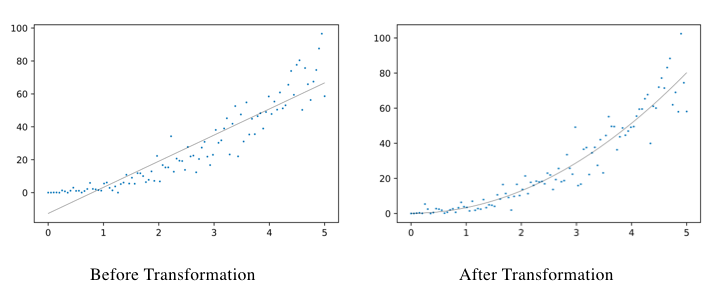}}
\caption{The effect of box-cox transformation in linear regression. The feature and label are quadratically related.}
\label{fig:boxcox}
\end{figure}

The transformation over dataset can help to obtain additional attributes. These features obtained by transformation could be unimportant for some data mining models, such as neural network, which have superior fitting potential. However, for relatively simpler models with fewer parameters, linear regression, for example, the transformed features do help the model to get better performance(as in \figref{fig:boxcox}), for they could provide additional indication of the relationship between attributes. The transformation would also be essential for scientific discoveries and machine controls\cite{lin2002attribute}.

Generally, given attributes set $\{a_1, a_2, \dots, a_p\}$, the numerical transformation can be represented as in \eqref{eq:general_trans}. Theoretically, $f$ can be any function, however, since the input data is finite, $f$ can take polynomial forms\cite{lin2002attribute}.

\begin{equation} \label{eq:general_trans}
    x' = f(a_1, a_2, \dots, a_p)
\end{equation}

The commonly used representation of $f$ includes \textit{polynomial based transformation}, \textit{approximation based transformation}, \textit{rank transformation} and \textit{box-cox transformation}\cite{garcia2015data}. The parameters in the transformation formula could be obtained by subjective definition (for situations where people know the relationship between attributes and labels well), by brute search\cite{garcia2015data} or by applying maximum likelihood method. 

\section{Data Reduction}

The amount of data in a data warehouse or a dataset can be huge, causing difficulties for data storage and processing when working on a data mining task, while not every model needs a huge amount of data to train. On the other hand, although the data may have lots of attributes, there could be unrelated features as well as the interdependence between features\cite{kotsiantis2006data}. Data reduction is the technique that helps to reduce the amount or dimension, or both, of a dataset, thus making the model's learning process more efficient as well as helping the model to obtain better performance, including preventing overfitting problem, and fix the skewed data distribution.

\subsection{Dimensional reduction}
The dimension reduction technique is about reducing the dimensionality of data samples, thus reduce the total size of the data. As the number of attributes is reduced for a sample, there is less information contained in it. A good dimensional reduction algorithm will keep more general information: this could make it more difficult for models to become overfitted.

\begin{figure}[!t]
\centering
\framebox{\includegraphics[width=2.5in]{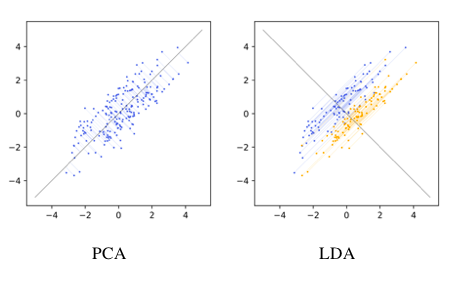}}
\caption{A comparison of reduction result with PCA and LDA.}
\label{fig:pcalda}
\end{figure}

Some dimensional reduction techniques pose a dimensional reduction transformation over a dataset, generating new data samples, which have a fewer number of attributes than before. The transformations have different criteria. Principal component analysis, known as PCA, could reduce the dimension of data while keeping the maximum variance of data\cite{jolliffe2011principal}. This is achieved by multiplying matrix $A = (a_1, a_2, \dots, a_p)^T$ to the dataset $X$ and keep the top $k$ dimensions ($a_i$ stands for the normalized eigenvector corresponds to the $i^{th}$ greatest eigenvalue of the covariance matrix of the dataset). By contrast, linear discriminant analysis (LDA), is meant to maximize the component axes for class-separation. The implementation of LDA is similar to PCA; the only difference is it replaces the covariance matrix with the scatter matrix of samples. A graphical illustration of the difference between PCA and LDA is given as \figref{fig:pcalda}. In comparatively more situations, LDA outperforms PCA. PCA may outperform LDA when the data amount is small, or the data is nonuniformly sampled\cite{martinez2001pca}. Other dimensional reduction algorithms include factor analysis (assuming a lower dimensional underlying distribution), projection pursuit(measuring the aspect of non-Gaussianity)\cite{fodor2002survey}, and wavelet transform\cite{han2011data}.

Feature selection is another dimensional reduction technique: it is about removing irrelevant or correlated attributes from the dataset while keeping the other relatively independent attributes untouched. Feature selection is more than simply selecting the feature that has greater relevance with the variable to predict, the relationship between attributes are also supposed to be taken into consideration: the goal is to find a sufficiently good subset of features to predict\cite{kotsiantis2006data}. Feature selection methods can be divided into three types, as follows\cite{guyon2003introduction}.

\begin{itemize}
    \item \textit{filter}: Directly select the feature based on attribute level criteria, including information gain, correlation score, or chi-square test. The filter method does not take the data mining model into consideration.
    \item \textit{wrapper}: Use techniques to search through the potential subsets, according to their performance on the data mining model. Greedy strategies, including forward selection and backward elimination\cite{guyon2003introduction}, are issued in order to reduce the time consumption.
    \item \textit{embedded}: Embed the feature selection into the data mining model. Usually, the weight over different attributes would act as feature selection. A typical example is the regularization term of the loss function in linear regression, known as Lasso regression (for L1 regularization) and ridge regression (for L2 regularization).
\end{itemize}

\subsection{Instance selection and sampling}
Both instance selection and sampling are about achieving the reduction of data by reducing the amount of data, seeking the chance to train the model with minimum performance loss\cite{garcia2015data}, while based on different criteria for selecting (or dropping) instance. 

Most instance selection algorithm is based on fine-tuning classification models. To help the model make better decision, \textit{condensation algorithm} and \textit{edition algorithm} are issued\cite{garcia2015data}. The condensation removes the samples lie in the relative center area of the class, assuming they do not contribute much in classification. \textit{Condensed nearest neighbor}\cite{hart1968condensed}, for instance, select instance by adding all the samples that cause a mistake to a K-Nearest Neighbor classifier. Edition algorithm removes the samples close to boundary, hoping to give the classifier a smoother decision boundary. Related algorithms include a clustering-based algorithm to select the center of clusters\cite{lumini2006clustering}, 

Compared with instance selection methods, sampling is a faster and easier way to reduce the number of instances, since almost no complex selection algorithm is required for sampling methods: they only focus on reducing the amount of the data samples. The easiest sampling technique is \textit{random sampling}, which collect a certain amount or portion of samples from the dataset randomly. For skewed datasets, \textit{stratified sampling}\cite{kotsiantis2006data} is more adapted, since it takes the appearance frequency of labels from different classes into account and assigns a different probability of data with different labels being chosen, thus makes the sampled dataset more balanced.

\section{Summary and Outlook}

Data mining, as a technique to discover additional information from a dataset, can be integrated as a pipeline, in which obtaining, scrubbing, exploring, modeling, and interpreting are the key steps. The purpose of the data mining pipeline is to tackle realistic problems, including reviewing the past and predicting the future. The specific technique used in each step should be selected with care to give the best performance to the pipeline.

The success of a data mining model depends on the proper data preprocessing work. The unpreprocessed data can be of unsuitable format for model input, causing instability for the optimization algorithm of the model, having a great impact on the model's performance because of its noise and outliers, and causing performance problems on the model's training process. With careful selection of preprocessing steps, these problems can be reduced or avoided.

Data type transformation techniques as well as missing value handling techniques makes it possible for models for processing different types of data. By applying normalization, the unit system of different attributes would be more unified, reducing the probability of an optimization algorithm to miss the global minimum. For simpler models, numerical transformation can provide richer features to the model, thus enhance the model's ability to discover more underlying relationships between features and labels. For the overfitting problems of the model, dimensional reduction techniques help model find the more general information about samples instead of the too detailed features by reducing the dimension of feature, thus remove some unimportant information. And for the performance of the model training, both dimensional reduction techniques and instance selection techniques would improve the training performance by reducing the total amount of data.

\ifCLASSOPTIONcaptionsoff
  \newpage
\fi

\bibliographystyle{IEEEtran}
\bibliography{bibliography}

\end{document}